\documentclass[runningheads]{llncs}
\usepackage{float}
\usepackage[caption = false]{subfig}
\usepackage{graphicx}
\usepackage{lipsum}
\usepackage{comment}
\usepackage{amsmath,amssymb} 
\usepackage{color}
\usepackage{enumitem}
\usepackage{subeqnarray}
\usepackage[ruled,vlined]{algorithm2e}
\usepackage{algpseudocode}
\usepackage{array,multirow}
\usepackage{float}

\begin{document}
\pagestyle{headings}
\mainmatter
\def\ECCVSubNumber{2983}  

\title{Few-Shot Scene-Adaptive Anomaly Detection} 

%
\author{Yiwei Lu\inst{1}\orcidID{0000-0001-7872-3186} \and
Frank Yu\inst{1}\orcidID{0000-0002-5620-8842 } \and
Mahesh Kumar Krishna Reddy\inst{1}\orcidID{0000-0001-5645-4931} \and Yang Wang\inst{1,2} \orcidID{0000-0001-9447-1791}
}
\authorrunning{Yiwei Lu et al.}
%
\institute{\textsuperscript{1} University of Manitoba, \textsuperscript{2}Huawei Technologies Canada \\
\email{\{luy2,kumark,ywang\}@cs.umanitoba.ca}}
\maketitle
\begin{abstract}
We address the problem of anomaly detection in videos. The goal is to identify unusual behaviours automatically by learning exclusively from normal videos. Most existing approaches are usually data-hungry and have limited generalization abilities. They usually need to be trained on a large number of videos from a target scene to achieve good results in that scene. In this paper, we propose a novel few-shot scene-adaptive anomaly detection problem to address the limitations of previous approaches. Our goal is to learn to detect anomalies in a previously unseen scene with only a few frames. A reliable solution for this new problem will have huge potential in real-world applications since it is expensive to collect a massive amount of data for each target scene. We propose a meta-learning based approach for solving this new problem; extensive experimental results demonstrate the effectiveness of our proposed method. All codes are released in \textcolor{blue}{\textsl{https://github.com/yiweilu3/Few-shot-Scene-adaptive-Anomaly-Detection}}.
\end{abstract}

\section{Introduction}
\begin{figure}[t]
    \centering
    \includegraphics[width= 0.7\textwidth]{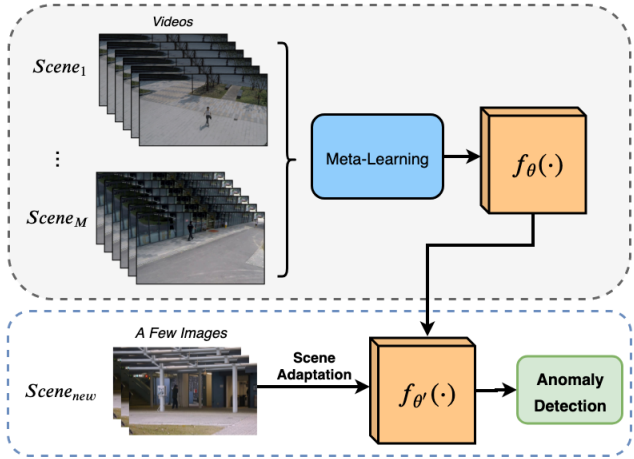}
\caption{An overview of our proposed problem setting. During training (1st row), we have access to videos collected from $M$ different camera scenes. From such training data, we use a meta-learning method to obtain a model $f_{\theta}$ with parameters $\theta$. Given a target scene (2nd row), we have access to a small number of frames from this target scene. Our goal is to produce a new model $f_{\theta'}$ where the model parameters $\theta'$ are specifically adapted to this scene. Then we can use $f_{\theta'}(\cdot)$ to perform anomaly detection on the remaining videos from this target scene.}
\label{fig:intro}
\end{figure}

We consider the problem of anomaly detection in surveillance videos. Given a video, the goal is to identify frames where abnormal events happen. This is a very challenging problem since the definition of ``anomaly'' is ambiguous -- any event that does not conform to ``normal'' behaviours can be considered as an anomaly. As a result, we cannot solve this problem via a standard classification framework since it is impossible to collect training data that cover all possible abnormal events. Existing literature usually addresses this problem by training a model using only normal data to learn a generic distribution for normal behaviours. During testing, the model classifies anomaly using the distance between the given sample and the learned distribution.

A lot of prior work~(e.g.~\cite{hasan2016learning,masci2011stacked,sabokrou2016video,chalapathy2017robust,sabokrou2018adversarially,abati2019latent,gong2019memorizing}) in anomaly detection use frame reconstruction. These approaches learn a model to reconstruct the normal training data and use the reconstruction error to identify anomalies. 
Alternatively, \cite{liu2018future,nguyen2019anomaly,lu2019future,luo2017remembering,medel2016anomaly} use future frame prediction for anomaly detection. These methods learn a model that takes a sequence of consecutive frames as the input and predicts the next frame. The difference between the predicted frame and the actual frame at the next time step is used to indicate the probability of an anomaly. 

However, existing anomaly detection approaches share common limitations. They implicitly assume that the model (frame reconstruction, or future frame prediction) learned from the training videos can be directly used in unseen test videos. This is a reasonable assumption only if training and testing videos are from the same scene (e.g. captured by the same camera). In the experiment section, we will demonstrate that if we learn an anomaly detection model from videos captured from one scene and directly test the model in a completely different scene, the performance will drop. Of course, one possible way of alleviating this problem is to train the anomaly detection model using videos collected from diverse scenes. Then the learned model will likely generalize to videos from new scenes. However, this approach is also not ideal. In order to learn a model that can generalize well to diverse scenes, the model requires a large capacity. In many real-world applications, the anomaly detection system is often deployed on edge devices with limited computing powers. As a result, even if we can train a huge model that generalizes well to different scenes, we may not be able to deploy this model. 

Our work is motivated by the following key observation. In real-world anomaly detection applications, we usually only need to consider one particular scene for testing since the surveillance cameras are normally installed at fixed locations. As long as a model works well in this particular scene, it does not matter at all whether the same model works on images from other scenes. In other words, we would like to have a model specifically adapted to the scene where the model is deployed. In this paper, we propose a novel problem called the \emph{few-shot scene-adaptive anomaly detection} illustrated in Fig.~\ref{fig:intro}. During training, we assume that we have access to videos collected from multiple scenes. During testing, the model is given a few frames in a video from a new target scene. Note that the learning algorithm does not see any images from the target scene during training. Our goal is to produce an anomaly detection model specifically adapted to this target scene using these few frames. We believe this new problem setting is closer to real-world applications. If we have a reliable solution to this problem, we only need a few frames from a target camera to produce an anomaly detection model that is specifically adapted to this camera. In this paper, we propose a meta-learning based approach to this problem. During training, we learn a model that can quickly adapt to a new scene by using only a few frames from it. This is accomplished by learning from a set of tasks, where each task mimics the few-shot scene-adaptive anomaly detection scenario using videos from an available scene.

This paper makes several contributions. First, we introduce a new problem called few-shot scene-adaptive anomaly detection, which is closer to the real-world deployment of anomaly detection systems. Second, we propose a novel meta-learning based approach for solving this problem. We demonstrate that our proposed approach significantly outperforms alternative methods on several benchmark datasets.

\section{Related Work}
\noindent {\bf Anomaly Detection in Videos}: Recent research in anomaly detection for surveillance videos can be categorized as either reconstruction-based or prediction-based methods. Reconstruction-based methods train a deep learning model to reconstruct the frames in a video and use the reconstruction error to differentiate the normal and abnormal events. Examples of reconstruction models include convolutional auto-encoders \cite{masci2011stacked,hasan2016learning,sabokrou2016video,chalapathy2017robust,gong2019memorizing}, latent autoregressive models~\cite{abati2019latent}, deep adversarial training \cite{sabokrou2018adversarially}, etc. Prediction-based detection methods define anomalies as anything that does not conform to the prediction of a deep learning model. Sequential models like Convolutional LSTM (ConvLSTM)~\cite{xingjian2015convolutional} have been widely used for future frame prediction and utilized to the task of anomaly detection \cite{luo2017remembering,medel2016anomaly}. Popular generative networks like generative adversarial networks (GANs)~\cite{goodfellow2014generative} and variational autoencoders (VAEs)~\cite{kingma2013auto} are also applied in prediction-based anomaly detection. Liu et al.~\cite{liu2018future} propose a conditional GAN based model with a low level optical flow \cite{dosovitskiy2015flownet} feature. Lu et al.~\cite{lu2019future} incorporate a sequential model in generative networks (VAEs) and propose a convolutional VRNN model. Moreover, \cite{gong2019memorizing} apply optical flow prediction constraint on a reconstruction based model. 

\noindent{\bf Few-Shot and Meta Learning}: To mimic the fast and flexible learning ability of humans, few-shot learning aims at adapting quickly to a new task with only a few training samples \cite{lake2015human}. In particular, meta learning (also known as \textsl{learning to learn}) has been shown to be an effective solution to the few-shot learning problem. The research in meta-learning can be categorized into three common approaches: metric-based \cite{koch2015siamese,vinyals2016matching,sung2018learning}, model-based \cite{santoro2016meta,munkhdalai2017meta} and optimization-based approaches \cite{ravi2016optimization,finn2017model}. Metric-based approaches typically apply Siamese \cite{koch2015siamese}, matching \cite{vinyals2016matching}, relation \cite{sung2018learning} or prototypical networks \cite{snell2017prototypical} for learning a metric or distance function over data points. Model-based approaches are devised for fast learning from the model architecture perspective \cite{santoro2016meta,munkhdalai2017meta}, where rapid parameter updating during training steps is usually achieved by the architecture itself. Lastly, optimization-based approaches modify the optimization algorithm for quick adaptation \cite{ravi2016optimization,finn2017model}. These methods can quickly adapt to a new task through the meta-update scheme among multiple tasks during parameter optimization. However, most of the approaches above are designed for simple tasks like image classification. In our proposed work, we follow a similar optimization-based meta-learning approach proposed in \cite{finn2017model} and apply it to the much more challenging task of anomaly detection. To the best of our knowledge, we are the first to cast anomaly detection as meta-learning from multiple scenes.

\section{Problem Setup}
\label{sec:set-up}

We first briefly summarize the standard anomaly detection framework. Then we describe our problem setup of \emph{few-shot scene-adaptive anomaly detection}.

\noindent {\bf Anomaly Detection:} The anomaly detection framework can be roughly categorized into reconstruction-based or prediction-based methods. For reconstruction-based methods, given a image $I$, the model $f_{\theta}(\cdot)$ generates a reconstructed image $\hat{I}$. For prediction-based methods, given $t$ consecutive frames $I_1,I_2,...,I_t$ in a video, the goal is to learn a model $f_{\theta}(x_{1:t})$ with parameters $\theta$ that takes these $t$ frames as its input and predicts the next frame at time $t+1$. We use $\hat{I}_{t+1}$ to denote the predicted frame at time $t+1$. The anomaly detection is determined by the difference between the predicted/reconstructed frame and the actual frame. If this difference is larger than a threshold, this frame is considered an anomaly.

During training, the goal is to learn the future frame prediction/reconstruction model $f_{\theta}(\cdot)$ from a collection of normal videos. Note that the training data only contain normal videos since it is usually difficult to collect training data with abnormal events for real-world applications. 

\noindent {\bf Few-Shot Scene-Adaptive Anomaly Detection:} The standard anomaly detection framework described above have some limitations that make it difficult to apply it in real-world scenarios. It implicitly assumes that the model $f_{\theta}(\cdot)$ (either reconstruction-based or prediction-based) learned from the training videos can generalize well on test videos. In practical applications, it is unrealistic to collect training videos from the target scene where the system will be deployed. In most cases, training and test videos will come from different scenes. The anomaly detection model $f_{\theta}(\cdot)$ can easily overfit to the particular training scene and will not generalize to a different scene during testing. We will empirically demonstrate this in the experiment section.

In this paper, we introduce a new problem setup that is closer to real-world applications. This setup is motivated by two crucial observations. First of all, in most anomaly detection applications, the test images come from a particular scene captured by the same camera. In this case, we only need the learned model to perform well on this particular scene. Second, although it is unrealistic to collect a large number of videos from the target scene, it is reasonable to assume that we will have access to a small number of images from the target scene. For example, when a surveillance camera is installed, there is often a calibration process. We can easily collect a few images from the target environment during this calibration process.

Motivated by these observations, we propose a problem setup called \emph{few-shot scene-adaptive anomaly detection}. During training, we have access to videos collected from different scenes. During testing, the videos will come from a target scene that never appears during training. Our model will learn to adapt to this target scene from only a few initial frames. The adapted model is expected to work well in the target scene.

\section{Our Approach: MAML for Scene-Adaptive Anomaly Detection}
\begin{figure*}[t]
    \centering
    \includegraphics[width= 1.0\textwidth]{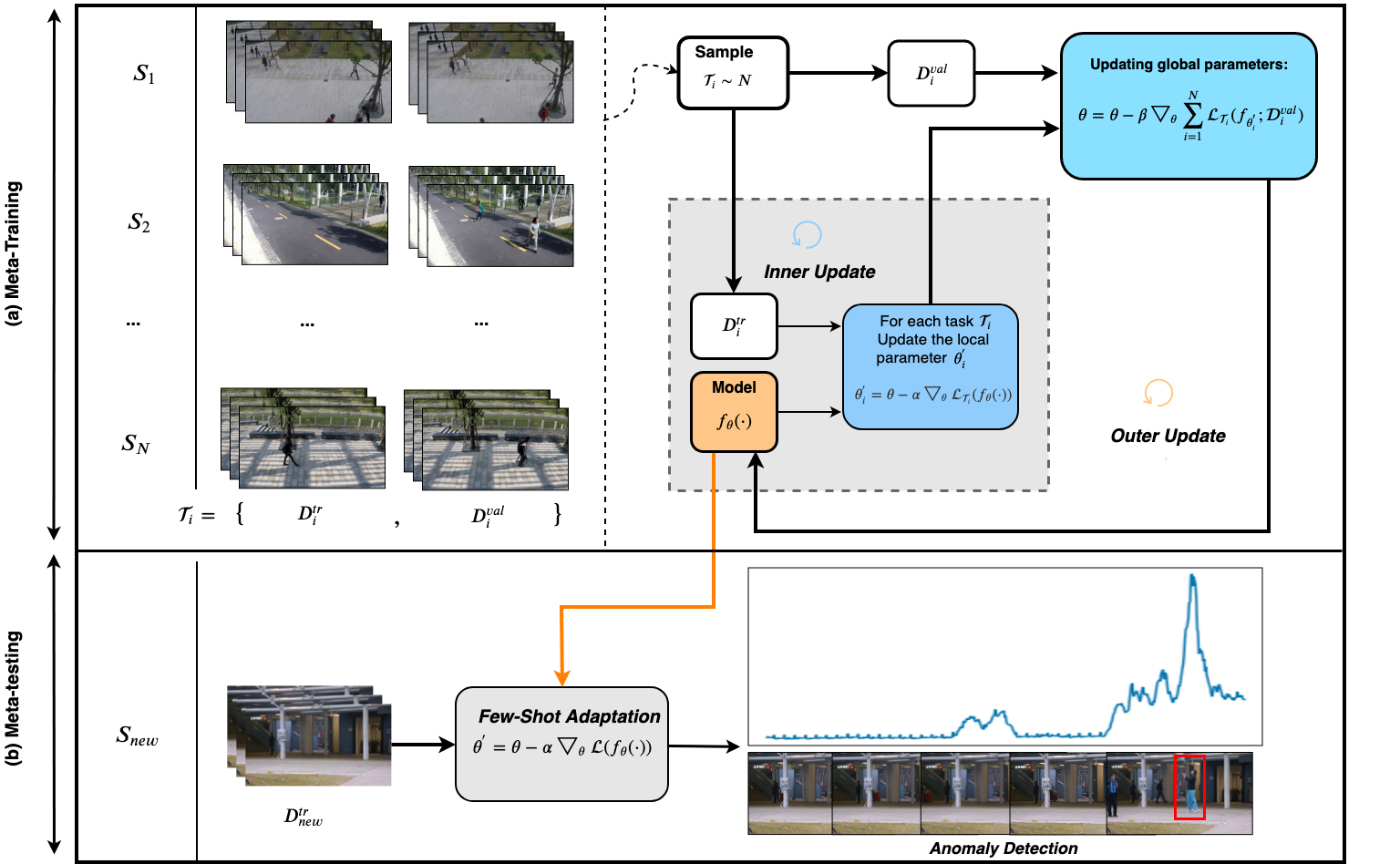}
    \caption{An overview of our proposed approach. Our approach involves two phases: (a) meta-training and (b) meta-testing. In each iteration of the meta-training (a), we first sample a batch of $N$ scenes $S_1, S_2,..., S_N$. We then construct a task $\mathcal{T}_i=\{D_i^{tr},D_i^{val}\}$ for each scene $S_i$ with a training set $D_i^{tr}$ and a validation set  $D_i^{val}$. $D_i^{tr}$ is used for \textit{inner update} through gradient descent to obtain the updated parameters $\theta'_{i}$ for each task. Then $D_i^{val}$ is used to measure the performance of $\theta'_{i}$. An \textit{outer update} procedure is used to update the model parameters $\theta$ by taking into account of all the sampled tasks. In meta-testing (b), given a new scene $S_{new}$, we use only a few frames to get the adapted parameters $\theta'$ for this specific scene. The adapted model is used for anomaly detection in other frames from this scene.}
    \label{fig:architecture}
\end{figure*}

We propose to learn few-shot scene-adaptive anomaly detection models using a meta-learning framework, in particular, the MAML algorithm~\cite{finn2017model} for meta-learning. Figure~\ref{fig:architecture} shows an overview of the proposed approach. The meta-learning framework consists of a meta-training phase and a meta-testing phase. During meta-training, we have access to videos collected from multiple scenes. The goal of meta-training is learning to quickly adapt to a new scene based on a few frames from it. During this phase, the model is trained from a large number of few-shot scene-adaptive anomaly detection tasks constructed using the videos available in meta-training, where each task corresponds to a particular scene. In each task, our method learns to adapt a pre-trained future frame prediction model using a few frames from the corresponding scene. The learning procedure (meta-learner) is designed in a way such that the adapted model will work well on other frames from the same scene. Through this meta-training process, the model will learn to effectively perform few-shot adaptation for a new scene. During meta-testing, given a few frames from a new target scene, the meta-learner is used to adapt a pre-trained model to this scene. Afterwards, the adapted model is expected to work well on other frames from this target scene.

Our proposed meta-learning framework can be used in conjunction with any anomaly detection model as the backbone architecture.  We first introduce the meta-learning approach for scene-adaptive anomaly detection in a general way that is independent of the particular choice of the backbone architecture, we then describe the details of the proposed backbone architectures used in this paper.

Our goal of few-shot scene-adaptive anomaly detection is to learn a model that can quickly adapt to a new scene using only a few examples from this scene. To accomplish this, the model is trained during a meta-training phase using a set of tasks where it learns to quickly adapt to a new task using only a few samples from the task. The key to applying meta-learning for our application is how to construct these tasks for the meta-training. Intuitively, we should construct these tasks so that they mimic the situation during testing. 

\noindent{\bf Tasks in Meta-learning}: 
We construct the tasks for meta-training as follows.\\
(1) Let us consider a future frame prediction model $f_{\theta}(I_{1:t})\rightarrow \hat{I}_{t+1}$ that maps $t$ observed frames $I_1,I_2,...,I_t$ to the predicted frame $\hat{I}_{t+1}$ at $t+1$. We have access to $M$ scenes during meta-training, denoted as $S_1,S_2,...,S_M$. For a given scene $S_i$, we can construct a corresponding task $\mathcal{T}_i = (\mathcal{D}_i^{tr},\mathcal{D}_i^{val})$, where $\mathcal{D}_i^{tr}$ and $\mathcal{D}_i^{val}$ are the training and the validation sets in the task $\mathcal{T}_i$. We first split videos from $S_i$ into many overlapping consecutive segments of length $t+1$. Let us consider a segment $(I_1,I_2,...,I_t,I_{t+1})$. We then consider the first $t$ frames as the input $x$ and the last frame as the output $y$, i.e. $x=(I_1,I_2,...,I_t)$ and $y=I_{t+1}$. This will form an input/output pair $(x,y)$. The future frame prediction model can be equivalently written as $f_{\theta}: x\rightarrow y$. In the training set $\mathcal{D}_i^{tr}$, we randomly sample $K$ input/output pairs from $\mathcal{T}_i$ to learn future frame prediction model, i.e. $\mathcal{D}^{tr}=\{(x_1,y_1), (x_2, y_2), ..., (x_K, y_K)\}$. Note that to match the testing scheme, we make sure that all the samples in $\mathcal{D}^{tr}$ come from the same video. We also randomly sample $K$ input/output pairs (excluding those in $\mathcal{D}_i^{tr}$) to form the test data $\mathcal{D}_i^{val}$.\\ 
(2) Similarly, for reconstruction-based models, we construct task  $\mathcal{T}_i = (\mathcal{D}_i^{tr},\mathcal{D}_i^{val})$ using individual frames. Since the groundtruth label for each image is itself, we randomly sample $K$ images from one video as $\mathcal{D}_i^{tr}$ and sample $K$ images from the same video as $\mathcal{D}_i^{val}$.

\noindent{\bf Meta-Training}: Let us consider a pre-trained anomaly detection model $f_{\theta}:x\rightarrow y$ with parameters $\theta$. Following MAML~\cite{finn2017model}, we adapt to a task $\mathcal{T}_i$ by defining a loss function on the training set $\mathcal{D}_i^{tr}$ of this task and use one gradient update to change the parameters from $\theta$ to $\theta'_i$:
\begin{subeqnarray}
  \label{eq:adapt}
&& \theta'_i= \theta - \alpha \bigtriangledown_{\theta} \mathcal{L}_{\mathcal{T}_i}(f_{\theta}; \mathcal{D}_{i}^{tr}), \textrm{ where}\\
&& \mathcal{L}_{\mathcal{T}_i}(f_{\theta}; \mathcal{D}_{i}^{tr})=\sum_{(x_j,y_j)\in\mathcal{D}_{i}^{tr}}L(f_{\theta}(x_j),y_j)
\end{subeqnarray}
where $\alpha$ is the step size. Here $L(f_{\theta}(x_j),y_j)$ measures the difference between the predicted frame $f_{\theta}(x_j)$ and the actual future frame $y_j$. We define $L(\cdot)$ by combine the least absolute deviation ($L_1$ loss) \cite{pollard1991asymptotics}, multi-scale structural similarity measurement ($L_{ssm}$ loss) \cite{wang2003multiscale} and gradient difference ($L_{gdl}$ loss) \cite{mathieu2015deep}:
\begin{equation}
L(f_{\theta}(x_j),y_j) = \lambda_1 L_1(f_{\theta}(x_j),y_j)+ \lambda_2 L_{ssm}(f_{\theta}(x_j),y_j)+ \lambda_3 L_{gdl}(f_{\theta}(x_j),y_j),
\end{equation}
where $\lambda_1, \lambda_2, \lambda_3$ are coefficients that weight between different terms of the loss function.

 The updated parameters $\theta'$ are specifically adapted to the task $\mathcal{T}_i$. Intuitively we would like $\theta'$ to perform on the validation set $\mathcal{D}_i^{val}$ of this task. We measure the performance of $\theta'$ on $\mathcal{D}_i^{val}$ as:
\begin{eqnarray}
  \mathcal{L}_{\mathcal{T}_i}(f_{\theta'}; \mathcal{D}_{i}^{val})=\sum_{(x_j,y_j)\in\mathcal{D}_{i}^{val}} L(f_{\theta'}(x_j), y_j)
  \label{eq:testloss}
\end{eqnarray}

The goal of meta-training is to learn the initial model parameters $\theta$, so that the scene-adapted parameters $\theta'$ obtained via Eq.~\ref{eq:adapt} will minimize the loss in Eq.~\ref{eq:testloss} across all tasks. Formally, the objective of meta-learning is defined as:
\begin{eqnarray}
  \min_{\theta}\sum_{i=1}^{M} \mathcal{L}_{\mathcal{T}_i}(f_{\theta'}; \mathcal{D}_{i}^{val})
  \label{eq:metaloss}
\end{eqnarray}
The loss in Eq.~\ref{eq:metaloss} involves summing over all tasks during meta-training. In practice, we sample a mini-batch of tasks in each iteration.  Algorithm~\ref{algo:metalearn} summarizes the entire learning algorithm. 
\begin{algorithm}
  \caption{Meta-training for few-shot scene-adaptive anomaly detection}
  \SetAlgoLined
  \KwIn{Hyper-parameters $\alpha, \beta$}
  Initialize $\theta$ with a pre-trained model $f_{\theta}(\cdot)$\;
\While {not done} {
Sample a batch of scenes $\{S_i\}_{i=1}^N$\;
\For {each $S_i$}{
Construct $\mathcal{T}_i=(\mathcal{D}^{tr}_i, \mathcal{D}^{val}_i)$ from $S_i$\;
Evaluate $\bigtriangledown_{\theta}\mathcal{L}_{\mathcal{T}_i}(f_{\theta}; \mathcal{D}_i^{tr})$ in Eq.~\ref{eq:adapt}\;
Compute scene-adaptative parameters $\theta'_i= \theta - \alpha \bigtriangledown_{\theta}\mathcal{L}_{\mathcal{T}_i}(f_{\theta}; \mathcal{D}_i^{tr})$\;
}
Update $\theta \leftarrow {\theta}-\beta \sum_{i=1}^{N}\bigtriangledown_{\theta} \mathcal{L}_{\mathcal{T}_i}(f_{\theta'_i}; \mathcal{D}_{i}^{val})$ using each $\mathcal{D}^{val}_i$ and $\mathcal{L}_{\mathcal{T}_i}$ in Eq.~\ref{eq:testloss}\;
}
\label{algo:metalearn}
\end{algorithm}

\setlength{\textfloatsep}{0pt}

\noindent{\bf Meta-Testing}: After meta-training, we obtain the learned model parameters $\theta$. During meta-testing, we are given a new target scene $S_{new}$. We simply use Eq.~\ref{eq:adapt} to obtain the adapted parameters $\theta'$ based on $K$ examples in $S_{new}$. Then we apply $\theta'$ on the remaining frames in the $S_{new}$ to measure the performance. We use the first several frames of one video in $S_{new}$ for adaptation and use the remaining frames for testing. This is similar to real-world settings where it is only possible to obtain the first several frames for a new camera.

\noindent{\bf Backbone Architecture:} Our scene-adaptive anomaly detection framework is general. In theory, we can use any anomaly detection network as the backbone architecture. In this paper, we propose a future frame prediction based backbone architecture similar to \cite{liu2018future}. Following \cite{liu2018future}, we build our model based on conditional GAN. One limitation of \cite{liu2018future} is that it requires additional low-level feature (ie. optical flows) and is not trained end-to-end. To capture spatial-temporal information of the videos, we propose to combine generative models and sequential modelling. Specifically, we build a model using ConvLSTM and adversarial training. This model consists of a generator and a discriminator. To build the generator, we apply a U-Net \cite{ronneberger2015u} to predict the future frame and pass the prediction to a ConvLSTM module \cite{xingjian2015convolutional} to retain the information of the previous steps. The generator and discriminator are adversarially trained. We call our model \textsl{r-GAN}. Since the backbone architecture is not the main focus of the paper, we skip the details and refers readers to the supplementary material for the detailed architecture of this backbone. In the experiment section, we will demonstrate that our backbone architecture outperforms \cite{liu2018future} even though we do not use optical flows.\\
We have also experiment with other variants of the backbone architecture. For example, we have tried using the ConvLSTM module in the latent space of an autoencoder. We call this variant \textsl{r-GAN*}. Another variant is to use a vartional autoencoder instead of GAN. We call this variant \textsl{r-VAE}. Readers are referred to the supplementary material for the details of these different variants. In the experiment, we will show that r-GAN achieves the best performance among all these different variants. So we use r-GAN as the backbone architecture in the meta learning framework.

\section{Experiments}
\label{sec:experiment}
In this section, we first introduce our datasets and experimental setup in Sec.~\ref{sec:expset}. We then describe some baseline approaches used for comparison in Sec.~\ref{sec:baselines}. Lastly, we show our experimental results and the ablation study results in Sec.~\ref{sec:results}. 

\subsection{Datasets and Setup}\label{sec:expset}

\begin{figure}[ht!]
	\begin{subfloat}
  \centering
  \setlength{\tabcolsep}{1pt}
  \renewcommand{\arraystretch}{0.5} 
  \begin{tabular}{ccccc}
    \includegraphics[width=2.35cm]{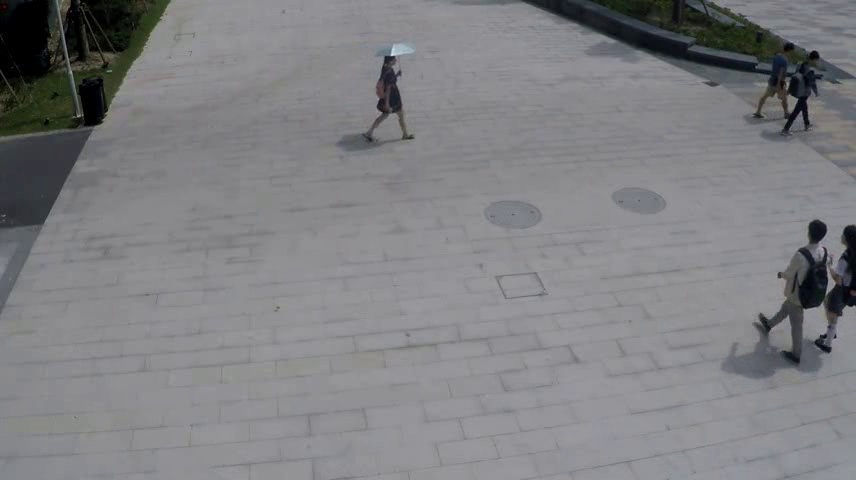}&\includegraphics[width=2.35cm]{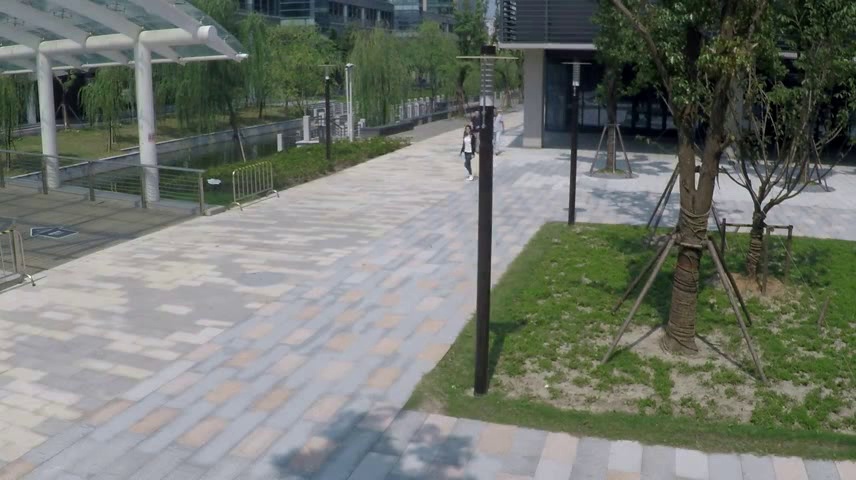}&\includegraphics[width=2.35cm]{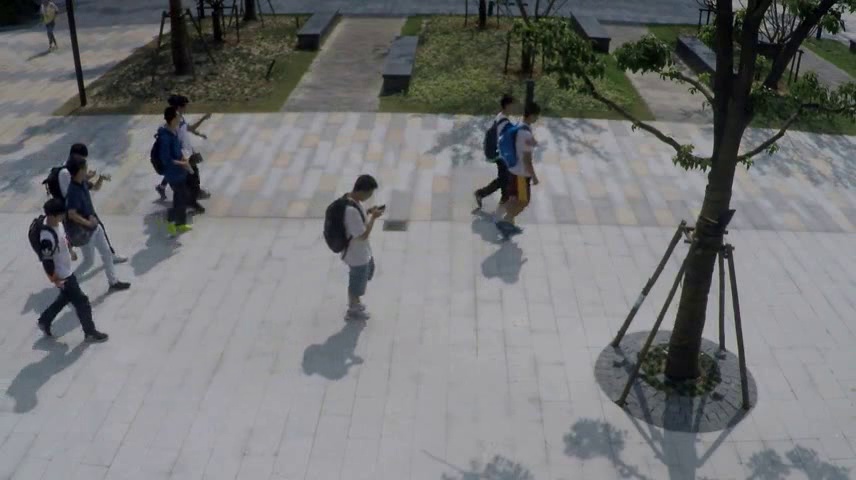}&\includegraphics[width=2.35cm]{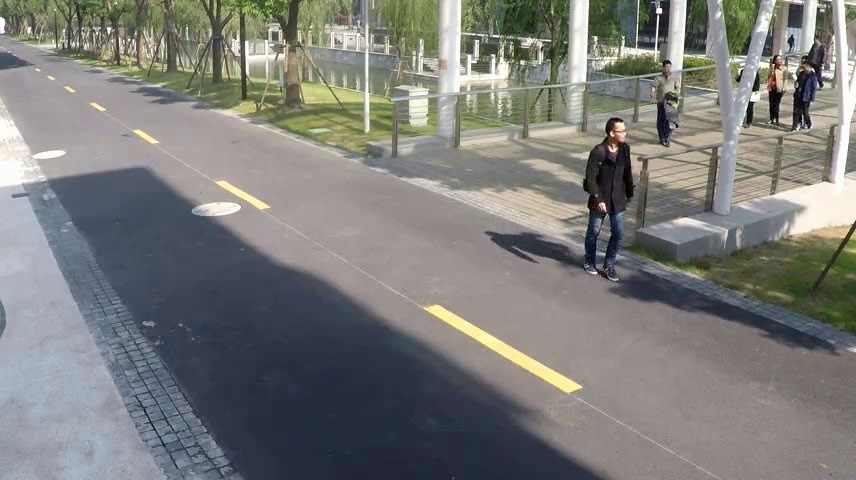}&\includegraphics[width=2.35cm]{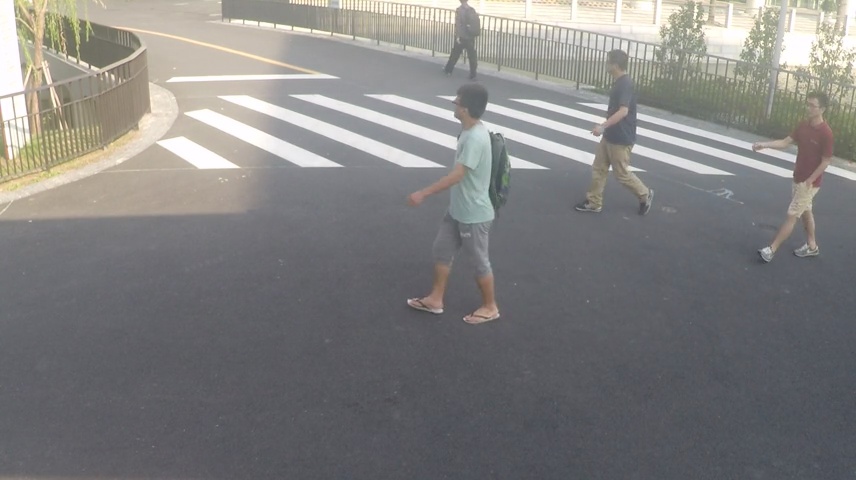}\\
    
    \includegraphics[width=2.35cm]{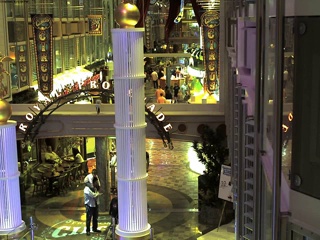}&\includegraphics[width=2.35cm]{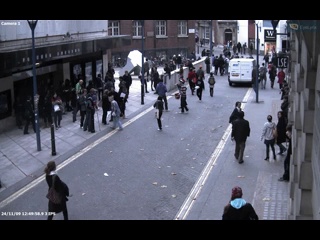}&\includegraphics[width=2.35cm]{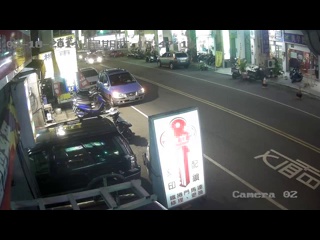}&\includegraphics[width=2.35cm]{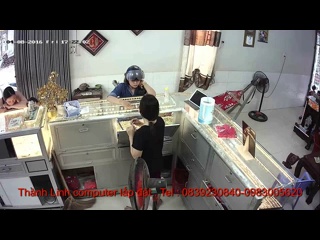}&\includegraphics[width=2.35cm]{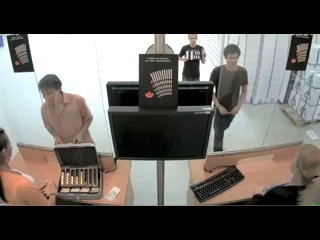}\\  
  \end{tabular}
  \caption{Example frames from the datasets used for meta-training. The first row shows examples of different scenes from the Shanghai Tech dataset. The second row shows examples of different scenes from the UCF crime dataset.}
  \label{fig: train}
\end{subfloat}
	\begin{subfloat}
	\centering
	\setlength{\tabcolsep}{1pt}
	\renewcommand{\arraystretch}{0.5} 
	\begin{tabular}{cccc}
	 \includegraphics[height=1.95cm]{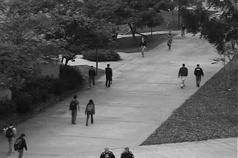}&\includegraphics[height=1.95cm]{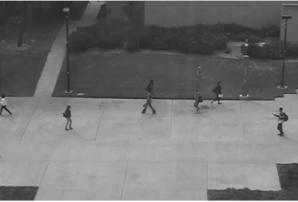}&\includegraphics[height=1.95cm]{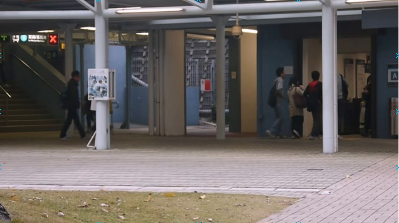}& \includegraphics[height=1.95cm]{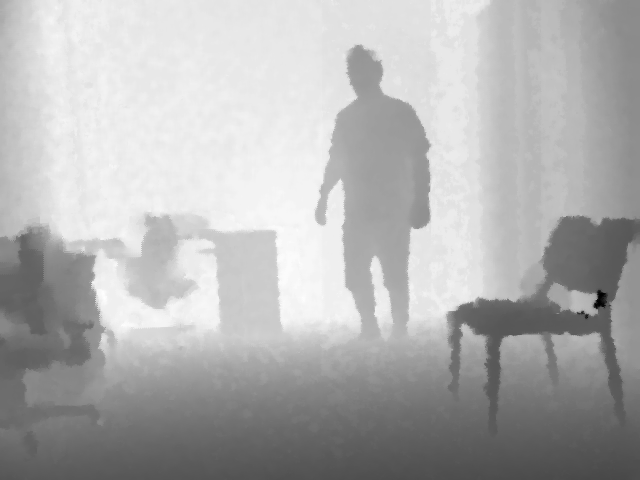}\\
	 \includegraphics[height=1.95cm]{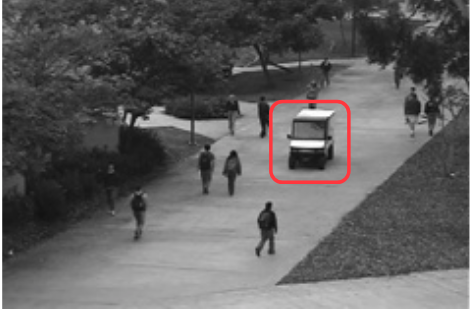}&\includegraphics[height=1.95cm]{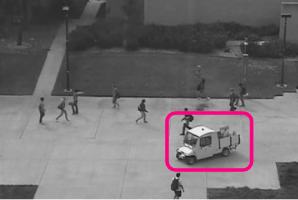}&\includegraphics[height=1.95cm]{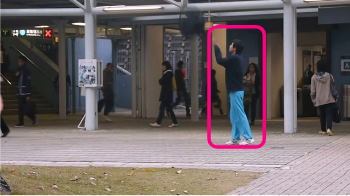}&\includegraphics[height=1.95cm]{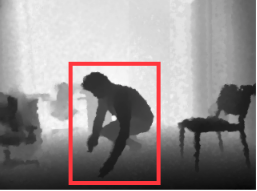} \\
	 Ped1&Ped2&Avenue&UR Fall
	\end{tabular}
	 \caption{Example frames from datasets used in meta-testing. The first row shows examples of normal frames for four datasets, and the second row shows the abnormal frames. Note that training videos only contain normal frames. Videos with abnormal frames are only used for testing.}
	 \label{fig: test}
\end{subfloat}
\end{figure}

\begin{table}[t!]
  \centering
  \small
  \begin{tabular}{|c|c|cccc|}
    \hline
     Category & Method &{Ped1}&{Ped2}&{CUHK}&{ST} \\ 
    \hline\hline
    \multirow{2}{*}{Feature} &
    MPCCA~\cite{kim2009observe} & 59.0 & 69.3 & - & - \\
    & Del et al.\cite{del2016discriminative} & - & - & 78.3 & - \\
     \cline{1-6}
    \multirow{5}{*}{Reconstruction} &
    Conv-AE \cite{hasan2016learning} & 75.0 & 85.0 & 80.0 & 60.9 \\
    & Unmasking \cite{tudor2017unmasking} & 68.4 & 82.2 & 80.6 & -\\  
    & LSA \cite{abati2019latent} & - & 95.4 & - & 72.5 \\
    & ConvLSTM-AE \cite{luo2017remembering} & 75.5 & 88.1 & 77.0 & -\\ 
     & MemAE \cite{gong2019memorizing} & - & 94.1 & 83.3 & 71.2 \\
     \cline{1-6}
     \multirow{5}{*}{Prediction} &
     Stacked RNN \cite{luo2017revisit} & - & 92.2 & 81.7 & 68.0   \\ 
    & FFP \cite{liu2018future} & 83.1 & 95.4 & 84.9 & 72.8  \\
    & MPED-RNN \cite{morais2019learning} & -& - & - & 73.4 \\ 
    & Conv-VRNN \cite{lu2019future} & \textbf{86.3} & 96.1 & 85.8 & - \\
    & Nguyen et al. \cite{nguyen2019anomaly} & - &
         \textbf{96.2} & \textbf{86.9} & - \\
    \cline{1-6}
    \multirow{3}{*}{Our backbones} &
    \textbf{r-VAE} & 82.4 & 89.2 & 81.8 & 72.7\\
    & \textbf{r-GAN*} & 83.7 & 95.9 & 85.3 & 73.7\\
   &  \textbf{r-GAN} & \textbf{86.3} & \textbf{96.2} & 85.8 & \textbf{77.9} \\
    \hline
\end{tabular}
\setlength{\abovecaptionskip}{15pt plus 3pt minus 2pt} 
\caption{Comparison of anomaly detection performance among our backbone architecture (r-GAN), its variants, and existing state-of-the-art in the standard setup (i.e. without scene adaptation). We report AUC~(\%) of different methods on UCSD Ped1 (Ped1), UCSD Ped2 (Ped2), CUHK Avenue (CUHK) and Shanghai Tech (ST) datasets. \textbf{We use the same train/test split as prior work on each dataset (i.e. without adaptation)}. Our proposed backbone architecture outperforms the existing state-of-the-art on almost all datasets.}
\label{tab:rGAN}
\end{table}

\begin{table}[t]
  \parbox{.45\linewidth}{
  \centering
  \small
  \begin{tabular}{|c|c|}
    \hline
    Method & AUC (\%)\\
    \hline\hline
    DAE~\cite{masci2011stacked} & 75.0\\
    CAE~\cite{masci2011stacked} & 76.0\\
    CLSTMAE \cite{nogas2019fall} & 82.0\\
    DSTCAE \cite{nogas2018deepfall} & 89.0\\
    \cline{1-2}
    \textbf{r-VAE} & 90.3 \\
    \textbf{r-GAN*} & 89.6\\
    \textbf{r-GAN} & {\bf 90.6}\\
    \hline
  \end{tabular}
    \setlength{\abovecaptionskip}{15pt plus 3pt minus 2pt}
\caption{Comparison of anomaly detection in terms of AUC~(\%) of different methods on the UR fall detection dataset. This dataset contains depth images. We simply treat those as RGB images. \textbf{We use the same train/test split as prior work on this dataset (i.e. without adaptation)}.Our proposed backbone architecture is state-of-the-art among all the methods.}
\label{tab:fall}
}
\hfill
  \parbox{.45\linewidth}{
  \centering
  \small
  \begin{tabular}{|c|ccc|}
    \hline
    Methods & $K=1$ & $K=5$ & $K=10$\\
    \hline\hline
    Pre-trained & 70.11 & 70.11 & 70.11\\ 
    Fine-tuned & 71.61 & 70.47 & 71.59\\
    {\bf Ours} & \textbf{74.51} & \textbf{75.28} & \textbf{77.36}\\ 
    \hline
  \end{tabular}
  \setlength{\abovecaptionskip}{15pt plus 3pt minus 2pt}
\caption{Comparison of $K$-shot scene-adaptive anomaly detection on the Shanghai Tech dataset. We use 6 scenes for training and the remaining 7 scenes for testing. We report results in terms of AUC~(\%) for $K=1,5,10$. The proposed approach outperforms two baselines.}
\label{tab:inner}
}
\end{table}

\noindent{\bf Datasets}: This paper addresses a new problem. In particular, the problem setup requires training videos from multiple scenes and test videos from different scenes. There are no existing datasets that we can directly use for this problem setup. Instead, we repurpose several available datasets.
\begin{itemize}[leftmargin=*]
\item Shanghai Tech~\cite{luo2017revisit}: This dataset contains 437 videos collected from 13 scenes. The training videos only contain normal events, while the test videos may contain anomalies. In the standard split in \cite{luo2017revisit}, both training and test sets contain videos from these 13 scenes. This split does not fit our problem setup where test scenes should be distinct from those in training. In our experiment, we propose a new train/test split more suitable for our problem. We also perform cross-dataset testing where we use the original Shanghai Tech dataset during meta-training and other datasets for meta-testing.
\item UCF crime~\cite{Sultani_2018_CVPR}: This dataset contains normal and crime videos collected from a large number of real-world surveillance cameras where each video comes from a different scene. Since this dataset does not come with ground-truth frame-level annotations, we cannot use it for testing since we do not have the ground-truth to calculate the evaluation metrics. Therefore, we only use the 950 normal videos from this dataset for meta-training, then test the model on other datasets. This dataset is much more challenging than Shanghai Tech when being used for meta-training, since the scenes are diverse and very dissimilar to our test sets. Our insight is that if our model can adapt to a target dataset by meta-training on UCF crime, our model can be trained with similar surveillance videos.
\item UCSD Pedestrian 1~\cite{mahadevan2010anomaly}, UCSD Pedestrian 2 (Ped 2)~\cite{mahadevan2010anomaly}, and CUHK Avenue~\cite{lu2013abnormal}: Each of these datasets contains videos from only one scene but different times. They contain 36, 12 and 21 test videos, respectively, including a total number of 99 abnormal events such as moving bicycles, vehicles, people throwing things, wandering and running. We use the model trained from Shanghai Tech or UCF crime datasets and test on these datasets.
\item UR fall \cite{kwolek2014human}: This dataset contains 70 depth videos collected with a Microsoft Kinect camera in a nursing home. Each frame is represented as a 1-channel grayscale image capturing the depth information. In our case, we convert each frame to an RGB image by duplicating the grayscale value among 3 color channels for every pixel. This dataset is originally collected for research in fall detection. We follow previous work in \cite{nogas2018deepfall} which considers a person falling as the anomaly. Again, we use this dataset for testing. Since this dataset is drastically different from other anomaly detection datasets, good performance on this dataset will be very strong evidence of the generalization power of our approach.
\end{itemize}
Figure~\ref{fig: train} and 
Figure~\ref{fig: test} show some example frames from the datasets we used in meta-training and meta-testing.

\noindent\textbf{Evaluation Metrics}: Following prior work~\cite{liu2018future,luo2017remembering,mahadevan2010anomaly}, we evaluate the performance using the area under the ROC curve (AUC). The ROC curve is obtained by varying the threshold for the anomaly score for each frame-wise prediction.

\noindent\textbf{Implementation Details}: We implement our model in PyTorch. We use a fixed learning rate of 0.0001 for pre-training. We fix the hyperparameters $\alpha$ and $\beta$ in meta-learning at 0.0001. During meta-training, we select the batch size of task/scenes in each epoch to be 5 on ShanghaiTech, and 10 on UCF crime.

\begin{table}[t!]
  \centering
  \small
  Shanghai Tech
  \begin{tabular*}{\textwidth}{c @{\extracolsep{\fill}}cccccccc}
    \hline
\textbf{Target} & \textbf{Methods} & \textbf{1-shot (K=1)} & \textbf{5-shot (K=5)} & \textbf{10-shot (K=10)}\\ 
\hline\hline
UCSD Ped 1 & Pre-trained & 73.1 & 73.1  & 73.1 \\ 
           & Fine-tuned & 76.99 & 77.85 & 78.23 \\ 
           & \textbf{Ours} & \textbf{80.6} & \textbf{81.42}  & \textbf{82.38} \\ 
\hline
UCSD Ped 2& Pre-trained & 81.95 & 81.95 & 81.95\\
          & Fine-tuned & 85.64 & 89.66 & 91.11\\
          & \textbf{Ours} & \textbf{91.19} & \textbf{91.8} & \textbf{92.8} \\
\hline
CUHK Avenue & Pre-trained & 71.43 & 71.43 & 71.43\\
            & Fine-tuned & 75.43 & 76.52 & 77.77\\ 
            & \textbf{Ours} & \textbf{76.58}  & \textbf{77.1}   & \textbf{78.79} \\
\hline
UR Fall & Pre-trained & 64.08 & 64.08 & 64.08 \\ 
        & Fine-tuned & 64.48 & 64.75 & 62.89\\
        & \textbf{Ours} & \textbf{75.51} & \textbf{78.7} & \textbf{83.24}\\
 \hline
\end{tabular*}
\end{table}
\begin{table}[t!]
  \centering
  \small
  UCF crime
  \begin{tabular*}{\textwidth}{c @{\extracolsep{\fill}}cccccccc}
    \hline
\textbf{Target} & \textbf{Methods} & \textbf{1-shot (K=1)} & \textbf{5-shot (K=5)} & \textbf{10-shot (K=10)}\\ 
\hline\hline
UCSD Ped 1 & Pre-trained & 66.87 & 66.87 & 66.87\\ 
           & Fine-tuned & 71.7 & 74.52 & 74.68\\ 
           & \textbf{Ours} & \textbf{78.44} & \textbf{81.43} & \textbf{81.62}\\ 
\hline
UCSD Ped 2& Pre-trained & 62.53 & 62.53 & 62.53\\
          & Fine-tuned & 65.58 & 72.63 & 78.32\\
          & \textbf{Ours}  & \textbf{83.08} & \textbf{86.41} & \textbf{90.21} \\
\hline
CUHK Avenue & Pre-trained & 64.32 & 64.32 & 64.32\\
            & Fine-tuned & 66.7 & 67.12 & 70.61\\ 
            & \textbf{Ours} & \textbf{72.62} & \textbf{74.68} & \textbf{79.02} \\
\hline
UR Fall & Pre-trained & 50.87 & 50.87 & 50.87\\ 
        & Fine-tuned & 57.02 & 58.08 & 62.82\\
        & \textbf{Ours} & \textbf{74.59} & \textbf{79.08} & \textbf{81.85} \\
 \hline
\end{tabular*}
\caption{Comparison of $K$-shot ($K=1,5,10$) scene-adaptive anomaly detection under the cross-dataset testing setting. We report results in terms of AUC~(\%) using the Shanghai Tech dataset and UCF crime dataset for meta-training. We compare our results with two baseline methods. Our results demonstrate the effectiveness of our method on few-shot scene-adaptive anomaly detection.}
\label{tab:result2}
\end{table}

\subsection{Baselines}
\label{sec:baselines}

To the best of our knowledge, this is the first work on the scene-adaptive anomaly detection problem. Therefore, there is no prior work that we can directly compare with. Nevertheless, we define the following baselines for comparison.

\noindent \textbf{Pre-trained}: This baseline learns the model from videos available during training, then directly applies the model in testing without any adaptation.

\noindent \textbf{Fine-tuned}: This baseline first learns a pre-trained model. Then it adapts to the target scene using the standard fine-tuning technique on the few frames from the target scene.

\subsection{Experimental Results}\label{sec:results}

\noindent{\bf Sanity Check on Backbone Architecture}: We first perform an experiment as a sanity check to show that our proposed backbone architecture is comparable to the state-of-the-art. Note that this sanity check uses the standard training/test setup (training set and testing set are provided by the original datasets), and our model can be directly compared with other existing methods. 
Table~\ref{tab:rGAN} shows the comparisons among our proposed architecture (r-GAN), its variants (r-GAN* and r-VAE), and other methods when using the standard anomaly detection training/test setup on several anomaly detection datasets. 
Table~\ref{tab:fall} shows the comparison on the fall detection dataset. We can see that our backbone architecture r-GAN outperforms its variants and the existing state-of-the-art methods on almost all the datasets. As a result, we use r-GAN as our backbone architecture to test our few-shot scene-adaptive anomaly detection algorithm in this paper.

\noindent\textbf{Results on Shanghai Tech}: In this experiment, we use Shanghai Tech for both training and testing. In the train/test split used in~\cite{liu2018future}, both training and test sets contain videos from the same set of 13 scenes. This split does not fit our problem. Instead, we propose a split where the training set contains videos of 6 scenes from the original training set, and the test set contains videos of the remaining 7 scenes from the original test set. This will allow us to demonstrate the generalization ability of the proposed meta-learning approach. Table \ref{tab:inner} shows the average AUC score over our test split of this dataset (7 scenes). Our model outperforms the two baselines. 

\begin{table}[t]
  \centering
  \small
  \begin{tabular}{|ccccc|}
\hline
\textbf{Target} & \textbf{Methods} & K=1 & K=5 & K=10\\
\hline\hline
Ped1 & Fine-tuned & 76.99 & 77.85 & 78.23\\ 
     & Ours ($N=1$) & 79.94 & 80.44 & 78.88\\ 
     & {\bf Ours ($N=5$)} &\textbf{80.6} & \textbf{81.42} & \textbf{82.38} \\ 
\hline
Ped2 & Fine-tuned & 85.64 & 89.66 & 91.11\\ 
     & Ours ($N=1$) & 90.73 & 91.5 & 91.11\\
     & {\bf Ours ($N=5$)} & \textbf{91.19} & \textbf{91.8} & \textbf{92.8}\\ 
\hline
CUHK & Fine-tuned & 75.43 & 76.52 & 77.77 \\ 
     & Ours ($N=1$) & 76.05 & 76.53 & 77.31 \\ 
     & {\bf Ours ($N=5$)} & \textbf{76.58} & \textbf{77.1} &\textbf{78.79}\\
\hline
\end{tabular}
  \setlength{\abovecaptionskip}{15pt plus 3pt minus 2pt}
\caption{Ablation study for using different number of sampled tasks ($N=1$ or $N=5$) during each epoch of meta-training. The results show that even the performance of training with one task is better than fine-tuning. However, a larger number of tasks is able to train an improved model.}
\label{tab:ablation}
\end{table}

\begin{figure*}[ht]
    \centering
    \includegraphics[width= 1.0\textwidth]{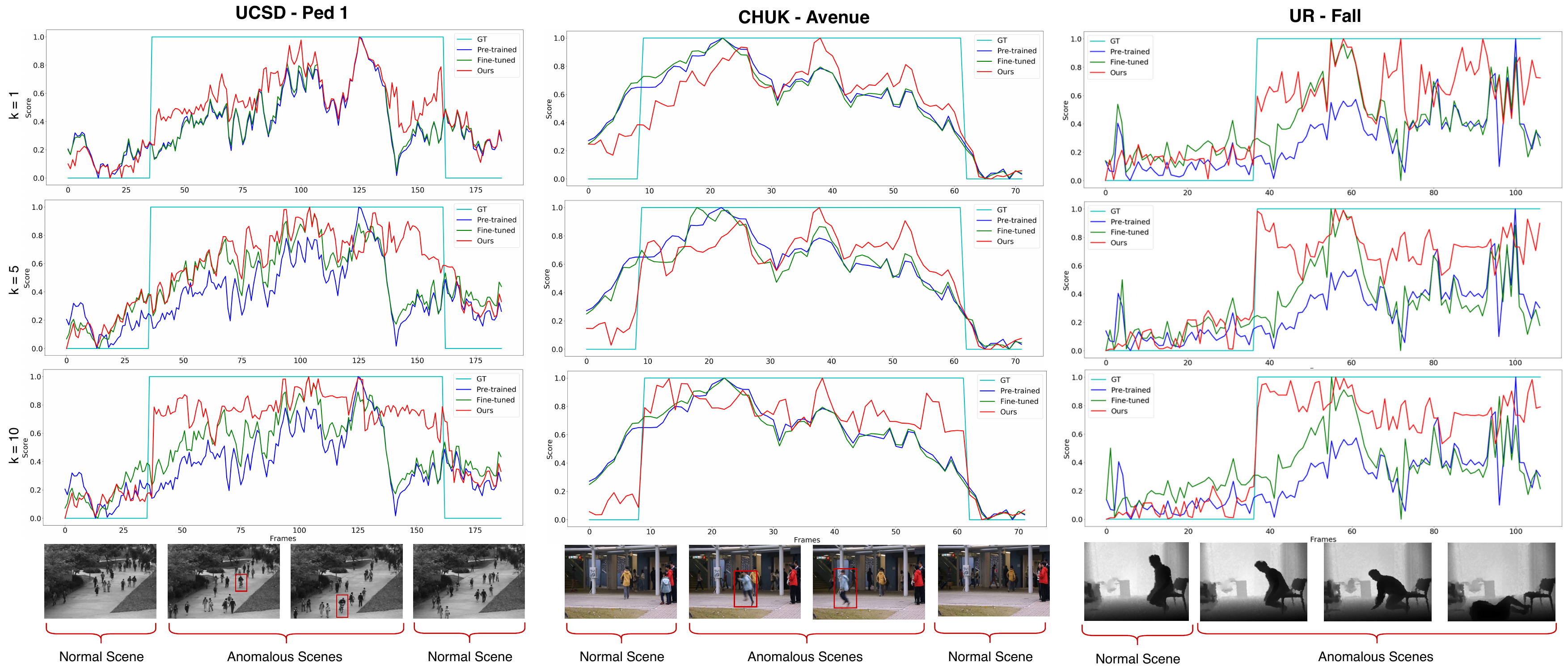}
    \caption{Qualitative results on three benchmark datasets using a pre-trained model on the Shanghai Tech dataset. Different columns represent results on different datasets. Each row shows few-shot scene-adaptive anomaly detection results with different numbers of training samples $K$. The red bounding boxes showing the abnormal event localization are for visualization purposes. They are not the outputs of our model which only predicts an anomaly score at the frame level.}
    \label{fig:qua}
\end{figure*}
\noindent\textbf{Cross-dataset Testing}: To demonstrate the generalization power of our approach, we also perform cross-dataset testing. In this experiment, we use either Shanghai Tech (the original training set) or UCF crime for meta-training, then use the other datasets (UCSD Ped1, UCSD Ped2, CUHK Avenue and UR Fall) for meta-testing. We present our cross-dataset testing results in Table \ref{tab:result2}. Compared with Table~\ref{tab:inner}, the improvement of our approach over the baselines in Table~\ref{tab:result2} is even more significant (e.g. more than $20\%$ in some cases). It is particularly exciting that our model can  successfully adapt to the UR Fall dataset, considering this dataset contains depth images and scenes that are drastically different from those used during meta-training.

\noindent{\bf Ablation Study}: In this study, we show the effect of the batch size (i.e. the number of sampled scenes) during the meta-training process. For this study, we train r-GAN on the Shanghai Tech dataset and test on Ped 1, Ped 2 and CUHK. We experiment with sampling either one ($N=1$) or five ($N=5$) tasks in each epoch during meta-training. Table~\ref{tab:ablation} shows the comparison. Overall, using our approach with $N=1$ performs better than simple fine-tuning, but not as good as $N=5$. One explanation is that by having access to multiple scenes in one epoch, the model is less likely to overfit to any specific scene.

\noindent{\bf Qualitative Results}: 
Figure~\ref{fig:qua} shows qualitative examples of detected anomalies. We visualize the anomaly scores on the frames in a video. We compare our method with the baselines in one graph for different values of $K$ and different datasets.

\section{Conclusion}
We have introduced a new problem called \emph{few-shot scene-adaptive anomaly detection}. Given a few frames captured from a new scene, our goal is to produce an anomaly detection model specifically adapted to this scene. We believe this new problem setup is closer to the real-world deployment of anomaly detection systems. We have developed a meta-learning based approach to this problem. During meta-training, we have access to videos from multiple scenes. We use these videos to construct a collection of tasks, where each task is a few-shot scene-adaptive anomaly detection task. Our model learns to effectively adapt to a new task with only a few frames from the corresponding scene. Experimental results show that our proposed approach significantly outperforms other alternative methods.

\noindent{\bf Acknowledgement:} This work was supported by the NSERC and UMGF funding. We thank NVIDIA for donating some of the GPUs used in this work.

\bibliographystyle{splncs04}
\bibliography{egbib}
\end{document}